\documentclass[10pt]{article}
\usepackage[T1]{fontenc}
\usepackage{times}
\usepackage[letterpaper,margin=0.92in]{geometry}
\usepackage{natbib}
\setcitestyle{authoryear,round,citesep={;},aysep={,},yysep={;}}
\usepackage{amsmath,amssymb,bm}
\usepackage{hyperref,url,booktabs,graphicx,array,tabularx,xcolor,colortbl}
\usepackage{microtype,enumitem,placeins,fvextra,caption,needspace,float}
\usepackage{fontawesome5}
\hypersetup{
    hidelinks,
    hypertexnames=false,
    pdftitle={LED-BENCH: A Benchmark for Large-format Engineering Drawing Understanding}
}
\graphicspath{{figures/}}
\newcommand{\ours}{LED-BENCH}
\newcolumntype{Y}{>{\raggedright\arraybackslash}X}
\renewcommand{\arraystretch}{1.16}
\title{
LED-BENCH: A Benchmark for Large-format
Engineering Drawing Understanding
}
\author{
{\fontsize{10.8}{13}\selectfont
Ke Li,\enspace
Tianyi Wang,\enspace
Enning You,\enspace
Chaoqun Wan,\enspace
Zhen Huang,\enspace
Shuxian Liang\textsuperscript{\ensuremath{\dagger}},\enspace
Xian-Sheng Hua
}\\[0.5em]
{\fontsize{10.2}{12.5}\selectfont
Institute of AI for Engineering, Tongji University
}\\[0.3em]
{\fontsize{10.2}{12.5}\selectfont
\href{https://iaie-physicalai.github.io/LED-BENCH}{%
\faGithub\enspace
\textcolor{blue!50!black}{https://iaie-physicalai.github.io/LED-BENCH}%
}
}
}
\date{}

\begin{document}
\maketitle
\begingroup
\renewcommand{\thefootnote}{\fnsymbol{footnote}}
\footnotetext[2]{Corresponding author.}
\endgroup
\vspace{0.4em}
\begin{center}
\includegraphics[
    width=\textwidth
]{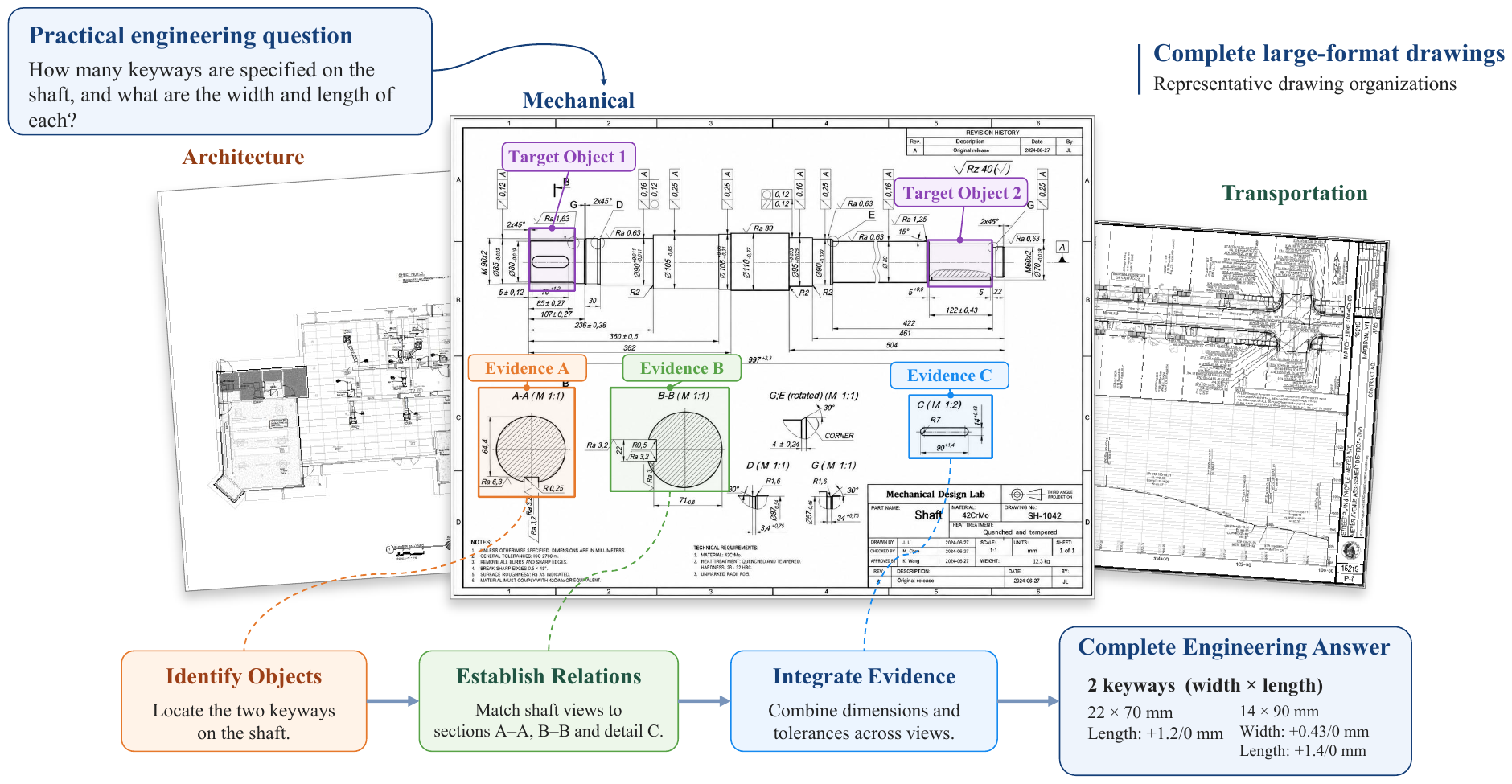}
\captionof{figure}{
    \textbf{Overview of LED-BENCH.}
    The schematic illustrates target localization,
    object association, and evidence integration
    over complete engineering drawings.
    Questions range from localized evidence
    understanding to cross-region or cross-view
    reasoning.
}
\label{fig:overview}
\end{center}
\vspace{0.5em}

\begin{abstract}
Real-world engineering drawings are often large-format and information-dense. Answering practical engineering questions therefore requires both fine-grained understanding of engineering information and complex reasoning over objects, relations, and evidence distributed across different regions or views. Existing benchmarks cover a range of engineering-drawing understanding tasks involving local regions, individual views, specific elements, or more constrained visual settings. However, systematic evaluation of understanding over realistic, large-format engineering drawings remains limited. To address this gap, we introduce LED-BENCH, a multidisciplinary benchmark designed to evaluate fine-grained understanding and complex reasoning over large-format engineering drawings, spanning the mechanical, architectural, and transportation domains. Extensive evaluation of 11 representative multimodal large language models, including 8 proprietary models and 3 open-source models, reveals substantial remaining difficulty: GPT-5.6 Sol achieves 52.65\% average accuracy, performance varies across engineering domains, and all evaluated models perform worse on questions requiring broader evidence integration. These results highlight fine-grained understanding, evidence integration, and complex reasoning over large-format engineering drawings as persistent challenges for current multimodal models.
\end{abstract}

\section{Introduction}
Engineering drawings communicate technical specifications through standardized graphical conventions used across mechanical engineering, architecture, and construction~\citep{iso2020technicaldrawings}. Beyond geometric shape, they carry dimensions, tolerances, materials, manufacturing requirements, and other information needed to define and realize a design~\citep{quintana2010mbd}. A complete large-format drawing may combine geometry, symbols, annotations, tables, notes, and information distributed across multiple views or distant regions. Answering an engineering question can therefore require reading fine-grained visual elements, associating them with the correct objects or views, reconstructing relations among engineering entities, and integrating evidence from different parts of the sheet. This setting combines the challenges of visually rich document understanding~\citep{mathew2021docvqa,mathew2022infographicvqa} with the loss of fine detail that can arise when multimodal models process high-resolution images~\citep{wang2025divide}.

Recent benchmarks establish important but complementary evaluation settings. MechVQA studies recognition, reasoning, and judging on mechanical drawings, while TriView2CAD evaluates orthographic-projection reasoning for CAD~\citep{kou2026mechvqa,niu2025creftcad}. DesignQA evaluates whether multimodal models can interpret and apply requirements from engineering documentation~\citep{doris2025designqa}. In architecture and construction, AECV-Bench evaluates object counting and drawing-grounded document question answering, ArchPlanVQA focuses on architectural floor-plan CAD drawings, and DrawingVQA organizes questions over real construction drawings by professional reasoning depth~\citep{kondratenko2026aecv,li2026archplanvqa,jung2026drawingvqa}. AEC-Bench further studies tool-using agents over construction document sets~\citep{mankodiya2026aecbench}. LED-BENCH provides a common evaluation setting for fine-grained understanding and complex reasoning over complete, large-format engineering drawings spanning mechanical, architectural, and transportation domains.

The benchmark is grounded in practical drawing-reading and engineering-analysis needs, covering both detailed local understanding and more complex questions requiring evidence association and integration across different regions or views. Mechanical drawings organize information through projections and detailed annotations, architectural drawings through spatial layouts, and transportation drawings through alignments and infrastructure networks. Evaluating these representations together provides a common basis for analyzing engineering drawing understanding across disciplines. 
Figure~\ref{fig:overview} illustrates the setting.

Table~\ref{tab:comparison} compares LED-BENCH with representative engineering-drawing benchmarks in terms of domain coverage, dataset scale, visual inputs, and task focus. In constructing LED-BENCH, engineering-background annotators define question requirements, review model-assisted drafts, and verify reference answers against the drawings.

\begin{table}[!htbp]
\centering

\caption{\textbf{Comparison with representative
engineering-oriented multimodal benchmarks.}}
\label{tab:comparison}

\fontsize{8.5pt}{10.5pt}\selectfont

\setlength{\tabcolsep}{3pt}
\renewcommand{\arraystretch}{1.10}
\renewcommand{\tabularxcolumn}[1]{m{#1}}

\begin{tabularx}{0.96\linewidth}{
@{\hspace{7pt}}
>{\raggedright\arraybackslash}m{2.65cm}
>{\raggedright\arraybackslash}m{1.40cm}
>{\centering\arraybackslash}m{1.10cm}
>{\centering\arraybackslash}m{1.15cm}
>{\raggedright\arraybackslash}m{2.85cm}
Y
@{\hspace{7pt}}
}

\toprule

\textbf{Benchmark} &
\textbf{Domain} &
\textbf{Images} &
\textbf{Tasks} &
\textbf{Visual Source} &
\textbf{Task Focus} \\

\midrule

MechVQA &
Mech. &
3,281 &
20,778 &
Mechanical drawings &
Recognition / Reasoning / Judging \\

\addlinespace[2pt]

AECV-Bench &
AEC &
21 &
192 &
AEC floor plans &
OCR / Counting / Spatial / Comparative \\

\addlinespace[2pt]

ArchPlanVQA &
Arch. &
11,477 &
26,266 &
Architectural floor plans &
Semantic understanding / Engineering reasoning \\

\addlinespace[2pt]

DrawingVQA &
AEC &
33 &
92 &
Construction drawings &
Perceptual / Contextual / Domain-expert \\

\addlinespace[2pt]

DesignQA &
Mech. &
230 &
1,451 &
Engineering drawings / CAD &
Requirement understanding / Compliance \\

\addlinespace[2pt]

TriView2CAD &
Civil / CAD &
203k &
$\approx$645K &
Bridge-pier three-view CAD &
Orthographic-projection reasoning \\

\addlinespace[2pt]

\rowcolor[HTML]{F1F5F8}

\textbf{LED-BENCH} &
\textbf{Mech. / Arch. / Transp.} &
\textbf{107} &
\textbf{377} &
\textbf{Large-format drawings} &
\textbf{Fine-grained understanding / Complex reasoning} \\

\bottomrule
\end{tabularx}

\end{table}

\FloatBarrier

Our evaluation covers 11 models over three independent
runs. Results reveal distinct domain-specific
performance patterns and consistently lower accuracy
on the \textit{Hard} subset. Supplementary experiments
quantify the gains obtained by providing additional
access to relevant drawing regions.

\Needspace{4\baselineskip}

Our main contributions are threefold:

\begin{enumerate}[
    leftmargin=16pt,
    itemsep=2pt,
    topsep=3pt,
    parsep=0pt,
    partopsep=0pt
]

    \item \textbf{A multidisciplinary benchmark for large-format engineering drawings.}
    We introduce 107 complete drawing sheets and 377 questions for evaluating fine-grained understanding and complex reasoning across mechanical, architectural, and transportation engineering.

    \item \textbf{An expert-led, model-assisted benchmark construction process.}
    We construct practical drawing-reading and engineering-analysis questions through an expert-led process in which model assistance supports candidate drafting and refinement, followed by drawing-grounded human verification.

    \item \textbf{A systematic evaluation of current multimodal models.}
    We evaluate 11 representative MLLMs under a unified three-run protocol, analyze performance across engineering domains and difficulty levels, and examine three visual-access interventions on the same question set.

\end{enumerate}

\section{Related Work}

\paragraph{Multimodal reasoning and visually rich document understanding.}
General-purpose multimodal benchmarks have progressively moved beyond object recognition toward knowledge-intensive and compositional reasoning. MMMU~\citep{yue2024mmmu} evaluates expert-level multimodal understanding across a broad range of academic disciplines, while MM-Vet~\citep{yu2024mmvet} examines the integration of multiple visual and linguistic capabilities. A closely related line of work studies visually rich documents, where answering questions depends not only on recognizing text but also on interpreting layout, graphical structure, and relations among multiple visual elements. DocVQA~\citep{mathew2021docvqa} established document-image question answering as a major evaluation setting; InfographicVQA~\citep{mathew2022infographicvqa} and ChartQA~\citep{masry2022chartqa} further require models to combine textual, graphical, and numerical evidence. OCRBench~\citep{liu2024ocrbench} systematically evaluates text-centric visual capabilities, while MMLongBench-Doc~\citep{ma2024mmlongbenchdoc} extends document reasoning to long-context settings in which relevant evidence may be distributed across multiple pages and heterogeneous document modalities.

Together, these studies show that multimodal document understanding requires substantially more than OCR. Engineering drawings share this need for localization and evidence integration, but introduce an additional representational layer: dimensions, projection conventions, technical symbols, geometric relations, stationing systems, and domain-specific drafting conventions jointly determine the meaning of the visual evidence.

\paragraph{Engineering drawing and design benchmarks.}
Recent benchmarks evaluate multimodal models directly on engineering representations, but differ substantially in domain, visual unit, and reasoning scope. MechVQA~\citep{kou2026mechvqa} studies large-scale mechanical-drawing understanding through fine-grained tasks spanning recognition, reasoning, and judging. DesignQA~\citep{doris2025designqa} evaluates the interpretation and application of engineering requirements using textual technical documents, CAD-related imagery, and engineering drawings. TriView2CAD, introduced alongside CReFT-CAD~\citep{niu2025creftcad}, focuses on orthographic-projection reasoning and correspondence between multi-view representations and underlying CAD or geometric structure.

In architecture and construction, AECV-Bench~\citep{kondratenko2026aecv} evaluates drawing-grounded capabilities including text interpretation, object counting, spatial reasoning, and comparison. ArchPlanVQA~\citep{li2026archplanvqa} studies visual question answering over architectural floor plans, while DrawingVQA~\citep{jung2026drawingvqa} emphasizes expert-curated reasoning questions grounded in real construction drawings. AEC-Bench~\citep{mankodiya2026aecbench} further extends evaluation toward agentic reasoning over real-world AEC document sets, spanning intra-sheet, intra-drawing, and intra-project tasks.

Taken together, prior benchmarks occupy complementary points along three axes: the engineering domain being represented, the visual unit supplied to the model, and the depth of reasoning required. \textit{\ours{}} focuses on their intersection in a complete-sheet setting, covering heterogeneous engineering domains and large-format drawings, with questions ranging from localized evidence understanding to cases in which information must be associated across different regions or views and integrated into a complete engineering answer.

\section{LED-BENCH}
\subsection{Drawing Collection}

LED-BENCH is constructed as a reasoning-intensive benchmark. We adopt a purposive, need-driven curation strategy focused on large-format engineering drawings that support fine-grained understanding and complex reasoning. Candidate drawings are collected from publicly accessible educational, technical, and project-document sources, including engineering textbooks and handbooks, public drawing repositories, and released design documents. Annotators inspect each candidate drawing and retain those that support meaningful drawing-reading or engineering-analysis questions involving detailed engineering information, relations among engineering objects, and, where applicable, evidence distributed across different regions or views. A drawing is retained only when it is sufficiently legible for reliable human interpretation, answer-relevant evidence can be grounded unambiguously, and it supports multiple distinct and determinate questions. Incomplete, corrupted, severely blurred, visually duplicated, or insufficiently legible drawings are excluded according to these curation criteria.

\subsection{Expert-Led Question Construction}

\begin{figure}[!htbp]
\centering
\includegraphics[width=\linewidth]{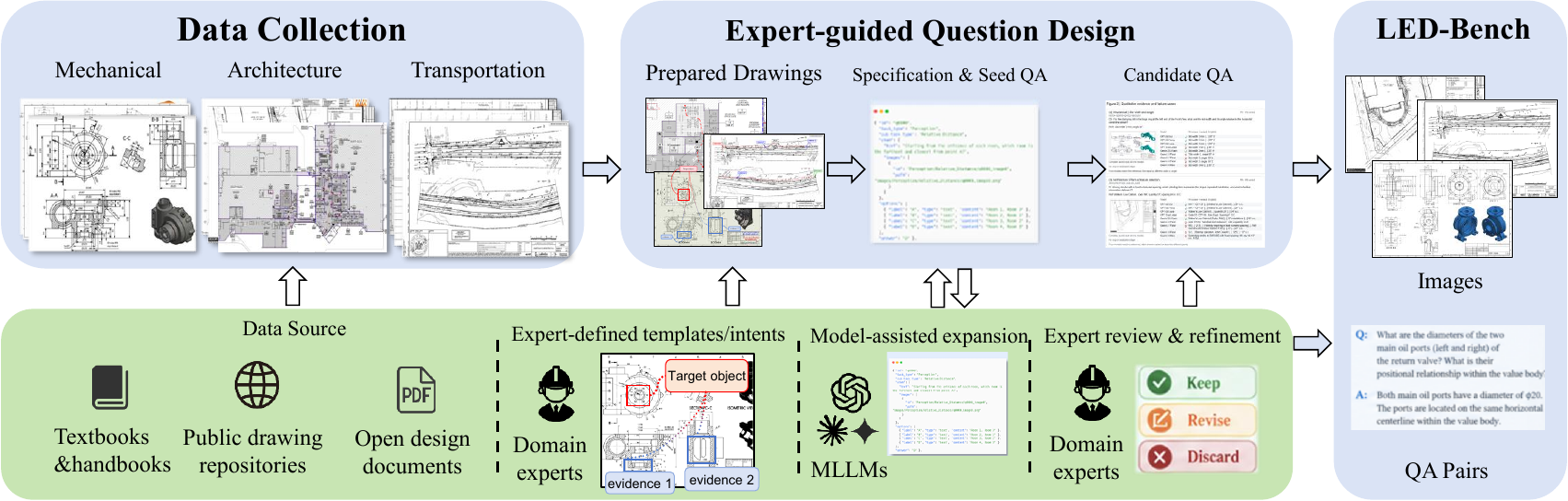}
\caption{\textbf{Expert-led, model-assisted benchmark construction.}
Domain experts specify questions and supporting evidence, language models expand the candidate set, and experts review each retained question and reference answer against its source drawing.}
\label{fig:construction}
\end{figure}

Figure~\ref{fig:construction} illustrates the expert-led, model-assisted construction process. For each drawing, engineering-background annotators identify a practical drawing-reading or engineering-analysis need and specify the target engineering object or task, the required evidence, the relations or reasoning involved, and the requirements for a complete answer.

Language models assist candidate drafting, answer proposals, and wording refinement. Engineering-background annotators verify each candidate and its reference answer against the corresponding source drawing, checking grounding, target-object association, required evidence, answer correctness and completeness, engineering relevance, ambiguity, and unintended answer cues. Candidates are revised or excluded when these criteria are not met. The retained questions are labeled \textit{Medium} or \textit{Hard} according to their evidence structure and reasoning requirements, as defined in Section~\ref{sec:dataset_overview}. Appendix~\ref{app:authoring} details the authoring process.

\subsection{Dataset Overview}
\label{sec:dataset_overview}
Table~\ref{tab:dataset_overview} summarizes the frozen benchmark. LED-BENCH contains 107 complete engineering drawing sheets and 377 questions: 45 mechanical drawings with 174 questions, 21 architectural drawings with 75 questions, and 41 transportation drawings with 128 questions. Multiple questions may be associated with the same source sheet, but each question is evaluated independently with a fresh model context.

\begin{table}[!htbp]
\centering
\caption{\textbf{Dataset composition across engineering domains.} Q/Draw. denotes the average number of questions associated with each complete drawing. Medium and Hard are relative difficulty levels within LED-Bench.}
\label{tab:dataset_overview}
\small
\setlength{\tabcolsep}{6.5pt}

\setlength{\tabcolsep}{7pt}
\begin{tabular}{@{\hspace{10pt}}lccccc@{\hspace{10pt}}}

\toprule
\textbf{Domain} & \textbf{Drawings} & \textbf{Questions} & \textbf{Q/Draw.} & \textbf{Medium} & \textbf{Hard} \\
\midrule
Mechanical     & 45 & 174 & 3.87 & 119 & 55 \\
Architecture   & 21 &  75 & 3.57 &  37 & 38 \\
Transportation & 41 & 128 & 3.12 &  55 & 73 \\
\midrule
\textbf{Total} & \textbf{107} & \textbf{377} & \textbf{3.52} & \textbf{211} & \textbf{166} \\
\bottomrule
\end{tabular}
\end{table}

\begin{figure}[!htbp]
\centering

\includegraphics[width=0.85\textwidth]{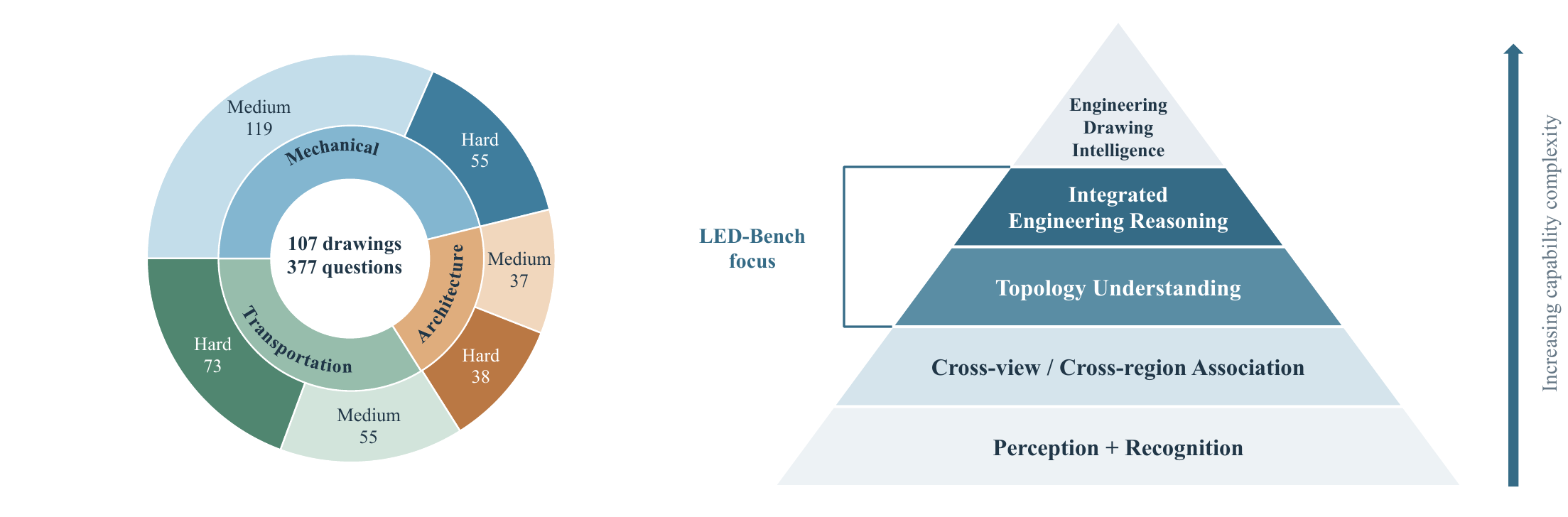}

\vspace{-1mm}
\begin{minipage}[t]{0.48\textwidth}
    \centering
    \textit{(a)} Dataset composition.
\end{minipage}
\hfill
\begin{minipage}[t]{0.48\textwidth}
    \centering
    \textit{(b)} Capability hierarchy.
\end{minipage}

\vspace{0.5mm}
\caption{\textbf{Dataset composition and benchmark focus.}
\textbf{(a)} Question distribution and \textit{Medium}/\textit{Hard} split across the three domains.
\textbf{(b)} Conceptual hierarchy of capabilities involved in large-format engineering drawing understanding and reasoning.}
\label{fig:dataset_stats}

\end{figure}

Figure~\ref{fig:dataset_stats} summarizes the domain composition of the benchmark and the capabilities targeted by its questions.

\paragraph{Difficulty levels.}
LED-BENCH uses two relative difficulty levels, \textit{Medium} and \textit{Hard}. \textit{Medium} questions typically rely on a bounded set of evidence and require operations such as combining several annotations, resolving a local structural or assembly relation, establishing a simple cross-region or cross-view correspondence, or performing a one-step engineering calculation. \textit{Hard} questions require broader or more interdependent evidence, such as associating information across different regions or multiple views, reconstructing spatial or topological relations, tracing ordered dependencies, performing multi-step inference or calculation, or jointly applying multiple engineering constraints. Both levels remain drawing-grounded reasoning tasks; the distinction reflects the scope and interdependence of evidence and reasoning rather than a separation between recognition and reasoning. Engineering-background experts review assignments against the corresponding drawings.

\paragraph{Drawing scale.}
LED-BENCH exhibits substantial variation in image scale across domains, with median long-side lengths of approximately 1,450 pixels for mechanical drawings, 6,800 pixels for architectural drawings, and 3,400 pixels for transportation drawings. This variation reflects the heterogeneous spatial organization of large-format engineering drawings. Appendix~\ref{app:resolution_ablation} further examines model sensitivity to input rasterization resolution.

\section{Evaluation Results}
\subsection{Evaluation Setup}
\label{sec:evaluation_protocol}

\paragraph{Testing models.}
We evaluate 11 representative multimodal large language models (MLLMs), including eight proprietary models and three open-weight models. The proprietary models are GPT-5.6 Sol, GPT-5.6 Luna, GPT-5.5, Gemini-3.5 Flash, Qwen3.8-Max, Qwen3.8-Flash, Qwen3.7-Flash, and Qwen3.6-Flash; the open-weight models are Kimi K2.5, Qwen3.6-27B, and Qwen3.8-27B.

\paragraph{Automatic scoring.}
LED-BENCH uses strict binary semantic scoring for open-ended answers. Each response is independently evaluated by three LLM judges using the question, reference answer, and required answer content, while the original drawing and evaluated-model identity are hidden from the judges. A response receives a score of 1 only when all indispensable engineering content is correct; semantically equivalent wording, numerical representations, and units are accepted, whereas substantive omissions, incorrect object or relation associations, numerical errors, or contradictions receive 0. Scores are aggregated from the judges' valid votes as detailed in Appendix~\ref{app:eval_config}. In a blinded review of 90 sampled responses, human labels agree with the LLM-majority decisions in 95.6\% of cases (Cohen's $\kappa=0.902$); Appendix~\ref{app:judge_validation} details this reference-based validation. Each model is evaluated in three independent runs over the full benchmark. Response scores are pooled across the runs within each domain, without majority voting across model responses. Because the three domains contain different numbers of questions, we use the unweighted domain average \(A_{\mathrm{avg}}=\frac{1}{3}(A_{\mathrm{mech}}+A_{\mathrm{arch}}+A_{\mathrm{trans}})\) as the primary metric.

Difficulty-level results are computed in the same way within the \textit{Medium} and \textit{Hard} subsets. Run-level results and stability statistics are reported in Appendix~\ref{app:run_stability}.

\subsection{Overall Performance}

Table~\ref{tab:main_results} reports performance across the three engineering domains. The highest domain-avg accuracy is 52.65\%, achieved by GPT-5.6 Sol, with 58.81\% on mechanical, 48.89\% on architectural, and 50.26\% on transportation drawings.

\begin{table}[!htbp]
\centering
\caption{\textbf{Performance across engineering domains.}
The best result in each column is shown in bold,
and the second-best is underlined.
Domain Avg uses unrounded domain scores;
recomputing from the two-decimal entries can
differ by 0.01 percentage points.}
\label{tab:main_results}

\small
\setlength{\tabcolsep}{8pt}
\renewcommand{\arraystretch}{1.12}

\begin{tabular}{@{\hspace{8pt}}lcccc@{\hspace{8pt}}}
\toprule
\textbf{Model} &
\textbf{Mechanical$\uparrow$} &
\textbf{Architecture$\uparrow$} &
\textbf{Transportation$\uparrow$} &
\textbf{Domain Avg$\uparrow$} \\
\midrule


\multicolumn{5}{c}{\textbf{Closed-source MLLMs}} \\
\addlinespace[1pt]
GPT-5.6 Sol          & \textbf{58.81} & \textbf{48.89} & \textbf{50.26} & \textbf{52.65} \\
GPT-5.6 Luna         & 50.38 & 47.11 & 44.27 & \underline{47.26} \\
Qwen3.8-Max          & 46.36 & \underline{48.00} & \underline{46.09} & 46.82 \\
Gemini-3.5 Flash     & \underline{57.47} & 35.11 & 31.77 & 41.45 \\
GPT-5.5              & 41.00 & 17.78 & 39.84 & 32.87 \\
Qwen3.7-Flash        & 32.57 & 16.44 & 27.86 & 25.63 \\
Qwen3.8-Flash        & 29.89 & 16.44 & 27.34 & 24.56 \\
Qwen3.6-Flash        & 18.39 & 6.22 & 15.36 & 13.33 \\
\midrule
\multicolumn{5}{c}{\textbf{Open-weight MLLMs}} \\
\addlinespace[1pt]
Kimi-K2.5             & 32.18 & 15.11 & 27.08 & 24.79 \\
Qwen3.8-27B           & 22.41 & 20.44 & 24.22 & 22.36 \\
Qwen3.6-27B           & 31.99 & 6.22 & 22.92 & 20.38 \\
\bottomrule
\end{tabular}
\end{table}

Performance also varies substantially across domains. For example, Gemini-3.5 Flash reaches 57.47\% on mechanical drawings but 35.11\% and 31.77\% on architectural and transportation drawings, respectively. In contrast, Qwen3.8-Max remains within a narrower range of 46.09--48.00\% across the three domains. These results indicate that aggregate performance alone can obscure substantially different domain-specific behavior.

\subsection{Performance across Difficulty Levels}

Table~\ref{tab:difficulty_results} compares performance on \textit{Medium} and \textit{Hard} questions using domain-avg accuracy. All evaluated models perform worse on the \textit{Hard} subset. GPT-5.6 Sol decreases from 56.57\% to 48.47\%, Qwen3.8-Max from 49.97\% to 42.09\%, and Gemini-3.5 Flash from 47.39\% to 34.49\%, the largest absolute drop among the evaluated models at 12.90 percentage points.

\begin{table}[!htbp]
\centering

\caption{\textbf{Domain-avg accuracy (\%) across the two difficulty levels.}
The best and second-best accuracies in each difficulty
level are shown in bold and underlined, respectively.
$\Delta$ denotes the absolute decrease from Medium to Hard.}
\label{tab:difficulty_results}

\small
\renewcommand{\arraystretch}{1.10}
\setlength{\tabcolsep}{7pt}

\begin{tabular*}{0.80\linewidth}{
@{\hspace{12pt}\extracolsep{\fill}}lccc@{\hspace{12pt}}
}
\toprule

\textbf{Model} &
\textbf{Medium$\uparrow$} &
\textbf{Hard$\uparrow$} &
\textbf{$\boldsymbol{\Delta}$} \\

\midrule

\multicolumn{4}{c}{\textbf{Closed-source MLLMs}} \\
\addlinespace[1pt]

GPT-5.6 Sol &
\textbf{56.57} &
\textbf{48.47} &
8.10 \\

GPT-5.6 Luna &
48.59 &
\underline{44.60} &
3.99 \\

Qwen3.8-Max &
\underline{49.97} &
42.09 &
7.88 \\

Gemini-3.5 Flash &
47.39 &
34.49 &
12.90 \\

GPT-5.5 &
34.34 &
29.36 &
4.98 \\

Qwen3.7-Flash &
27.01 &
23.91 &
3.10 \\

Qwen3.8-Flash &
27.21 &
20.63 &
6.58 \\

Qwen3.6-Flash &
16.94 &
8.57 &
8.37 \\

\midrule

\multicolumn{4}{c}{\textbf{Open-weight MLLMs}} \\
\addlinespace[1pt]

Kimi-K2.5 &
25.98 &
22.68 &
3.30 \\

Qwen3.8-27B &
25.80 &
18.04 &
7.76 \\

Qwen3.6-27B &
23.59 &
16.86 &
6.73 \\

\bottomrule
\end{tabular*}

\end{table}

The accuracy gap is consistent with the intended \textit{Medium}/\textit{Hard} distinction: \textit{Hard} questions generally require broader or more interdependent evidence and more involved reasoning.

\subsection{Failure Stages and Diagnostic Interventions}
\label{sec:error_analysis}

\begin{figure}[H]
\centering
\includegraphics[width=0.8\textwidth]{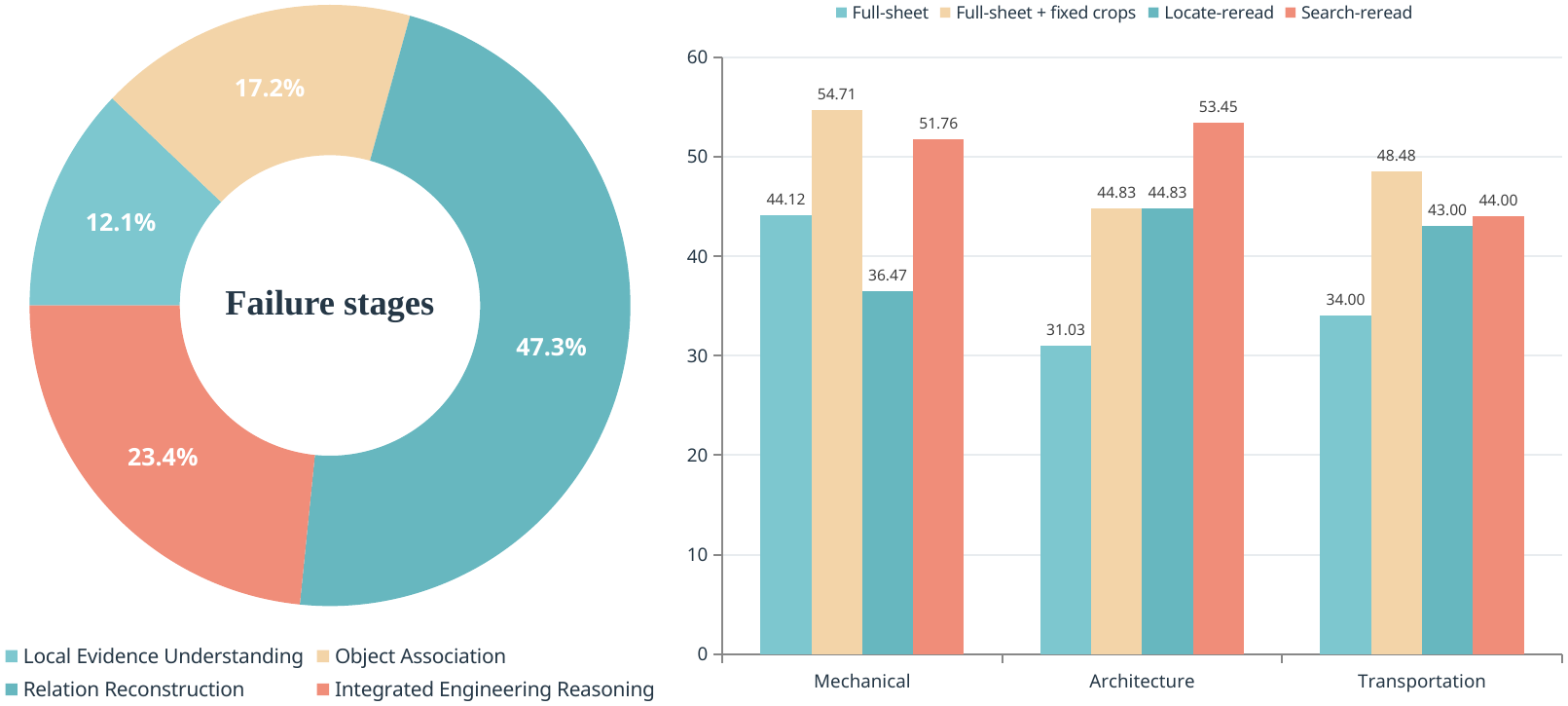}
\vspace{-1.5mm}
\begin{minipage}[t]{0.48\textwidth}
    \centering
    \textbf{(a)} Failure-stage distribution
\end{minipage}
\hfill
\begin{minipage}[t]{0.48\textwidth}
    \centering
    \textbf{(b)} Accuracy across domains
\end{minipage}
\vspace{0.5mm}
\caption{\textbf{Failure stages and domain-level intervention performance.} \textbf{(a)} Distribution of the four failure stages in the coded error sample used for the diagnostic analysis. \textbf{(b)} Four-model mean accuracy across Mechanical, Architecture, and Transportation for whole-sheet direct answering and three visual-access interventions.}
\label{fig:error_intervention}
\end{figure}

We analyze model failures at four conceptual stages: \textit{Local Evidence Understanding}, \textit{Object Association}, \textit{Relation Reconstruction}, and \textit{Integrated Engineering Reasoning}. In the coded error sample shown in Figure~\ref{fig:error_intervention}, \textit{Relation Reconstruction} is the largest category (47.3\%), followed by \textit{Integrated Engineering Reasoning} (23.4\%), \textit{Object Association} (17.2\%), and \textit{Local Evidence Understanding} (12.1\%).

To examine whether improved visual access can alleviate these failures, we conduct a supplementary diagnostic study on 377 questions from 107 drawings, comprising 174 mechanical, 75 architectural, and 128 transportation questions. Four response models---GPT-5.4, Qwen3.8-Flash, Gemini-3.5 Flash, and GPT-5.6 Sol---are evaluated for three runs under a whole-sheet baseline and three visual-access interventions. The \textit{fixed overlapping crops} setting supplements the full drawing with four predefined overlapping crops. \textit{Locate then reread} first identifies likely target regions and then rereads the corresponding crops. \textit{Search then reread} searches for multiple relevant regions before rereading and integrating the resulting observations.

\begin{table}[H]
\centering

\captionsetup{font=footnotesize}
\caption{Domain-avg accuracy (\%) for the whole-sheet baseline and three visual-access interventions. $\Delta$ denotes the absolute change relative to whole-sheet direct answering.}
\label{tab:error_intervention}

\small
\renewcommand{\arraystretch}{1.10}
\setlength{\tabcolsep}{9pt}

\begin{tabular}{@{\hspace{12pt}}llrr@{\hspace{12pt}}}
\toprule

\textbf{Method} &
\textbf{Intended support} &
\textbf{Domain Avg$\uparrow$} &
\textbf{$\Delta$} \\

\midrule

Whole-sheet direct answering &
Baseline &
36.38 & -- \\

Fixed overlapping crops &
Local evidence access &
49.34 & 12.96 \\

Locate then reread &
Target localization + rereading &
41.43 & 5.05 \\

Search then reread &
Multi-region search + rereading &
\textbf{49.74} & 13.35 \\

\bottomrule
\end{tabular}

\end{table}
\FloatBarrier
\Needspace{6\baselineskip}

The three interventions exhibit different cross-domain behavior. Fixed overlapping crops improve accuracy in all three domains and raise the domain-avg score from 36.38\% to 49.34\%. Search then reread reaches a similar overall level (49.74\%) and gives the highest architectural accuracy (53.45\%). Locate then reread is more mixed: it improves Architecture and Transportation but decreases Mechanical from 44.12\% to 36.47\%. Relative to whole-sheet direct answering, the visual-access interventions achieve higher overall accuracy, with the largest gains differing across domains and methods. Their input procedures and model-call configurations are detailed in Appendix~\ref{app:interventions}.

\section{Conclusion}
LED-BENCH establishes a multidisciplinary benchmark for fine-grained understanding and complex reasoning over complete engineering drawings across mechanical, architectural, and transportation domains. Under a unified three-run evaluation protocol, the 11 evaluated MLLMs show substantial cross-domain variation and consistently lower accuracy on questions requiring broader and more interdependent evidence. Supplementary visual-access interventions improve performance in several settings, with different effects across methods and domains. These findings highlight the importance of locating, associating, and integrating engineering evidence across complete drawing sheets.

\clearpage
\appendix

\Needspace{34\baselineskip}
\section{Drawing Scale and Visual Input Considerations}
\label{app:drawing_scale}
The three domains differ substantially in raster dimensions, reflecting both their heterogeneous source types and drawing conventions. For native mechanical rasters and the frozen 200-DPI architectural/transportation renders, the median long-side lengths are approximately 1,450 pixels for mechanical drawings, 6,800 pixels for architectural drawings, and 3,400 pixels for transportation drawings. Mechanical sheets often contain dense multi-view geometry and technical annotations, while architectural and transportation sheets distribute evidence across larger spatial extents.

These dimensions describe images at submission. Model APIs apply their own visual preprocessing, including resizing, tiling, and visual tokenization.

\subsection{Exploratory Rasterization Results}
\label{app:resolution_ablation}
\begin{figure}[!htbp]
\centering
\includegraphics[width=0.78\linewidth]{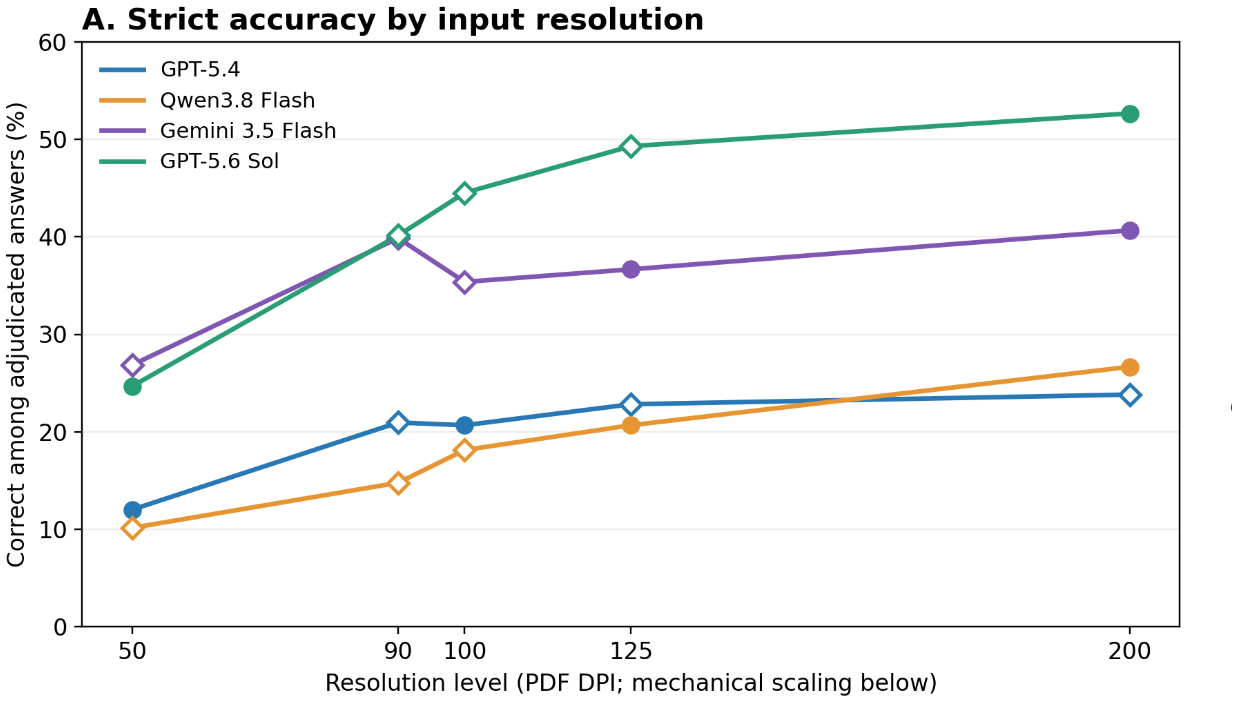}
\caption{\textbf{Exploratory rasterization study.} Strict accuracy among adjudicated answers as a function of client-side PDF rasterization resolution for four representative response models. Results are reported across the tested rendering resolutions.}
\label{fig:resolution_ablation}
\end{figure}

Figure~\ref{fig:resolution_ablation} summarizes an exploratory rasterization study on architectural and transportation PDF-source drawings. We compare multiple client-side rendering resolutions and report the strict accuracy among adjudicated answers for four response models. Higher input resolution generally improves accuracy, particularly for GPT-5.6 Sol and Gemini 3.5 Flash, with variation across models and rendering levels.

The main benchmark uses 200 DPI for PDF-source drawings and preserves native mechanical raster images. The exploratory study characterizes how client-side rendering resolution affects model performance.

\FloatBarrier

\section{Evaluation Protocol Details}
\label{app:eval_config}
Each question receives a fresh request with the complete image and no retrieval tools, demonstrations, or reference answer. The inference client preserves the submitted image bytes.

\begin{sloppypar}Its wrapper sends \nolinkurl{max_completion_tokens=32768} and normally requests \nolinkurl{response_format=json_object}. Temperature, top-$p$, seed, reasoning effort, and image-detail parameters are omitted from the requests; their effective values are provider defaults. The client requests a short final answer, and an interface-error fallback can remove the JSON response-format constraint. The completion-token limit applies to provider-accounted output tokens, including reasoning tokens where applicable.\end{sloppypar}

R1--R3 use the same request template with independent calls and response-level scoring. Initial pass-level deadlines are 240/600/600 seconds, with deferred technical retries; recovery schedules and attempt metadata are recorded in the run artifacts. Successful responses are retained, and retries address technical failures. Per-attempt records contain parsing status, returned model IDs, finish reasons, and retry information.

\begin{sloppypar}The eleven displayed main-table API aliases are \nolinkurl{gpt-5.6-sol}, \nolinkurl{gpt-5.6-luna}, \nolinkurl{gpt-chat-latest}, \nolinkurl{gemini-3.5-flash}, \nolinkurl{qwen3.8-max}, \nolinkurl{qwen3.8-flash}, \nolinkurl{qwen3.7-flash}, \nolinkurl{qwen3.6-flash}, \nolinkurl{kimi-k2.5}, \nolinkurl{qwen3.8-27b}, and \nolinkurl{qwen3.6-27b}. The gateway alias \nolinkurl{gpt-chat-latest} is reported as GPT-5.5.\end{sloppypar}

\subsection{Exact answering prompt}
\begin{Verbatim}
You are an expert engineering-drawing reader. Use only evidence visible in the attached complete drawing.
\end{Verbatim}
\begin{Verbatim}
Discipline: {domain_label}.
Answer only from evidence visible in the attached complete drawing.
Do not use external standards or assumptions to fill missing information.
Do not describe your reasoning. Preserve identifiers, specifications, numbers, units, tolerances, directions, object-value bindings, and necessary order.
Answer in the same language as the question. Include every necessary answer field.

Question: {question}

Return exactly one JSON object with no Markdown:
{"answer":"...","answerability":"answerable|partial|insufficient|ambiguous","confidence":0.0}
\end{Verbatim}
The parser accepts a JSON object, including one within a code fence, and can recover a decodable object containing an \texttt{answer} field. Empty, truncated, or unusable responses are retained as technical or response failures. Confidence and answerability fields are stored but do not weight correctness.

\subsection{Exact binary judge prompt}
Each judge receives only the question, full reference answer, mandatory components, and final candidate answer. Candidate-model identity and the source image are absent. The component list contains the full reference answer as one item, and judges return an integer score of 0 or 1 with a brief explanation.
\begin{Verbatim}
You are a strict evaluator for an engineering-drawing question-answering benchmark. Judge only whether the candidate answer is fully correct with respect to the provided reference answer and mandatory components. Do not infer the candidate model identity. Return JSON only.
\end{Verbatim}
\begin{Verbatim}
Judge whether the candidate answer is fully correct.

Question:
{question}

Reference answer:
{reference_answer}

Mandatory components:
{mandatory_components}

Candidate answer:
{candidate_answer}

Rules:
- Score 1 only if the candidate is semantically correct and ALL mandatory components are satisfied.
- Score 0 if any mandatory component is missing or wrong.
- Score 0 for a wrong engineering object/instance, swapped values, wrong relation/direction/identifier, or a material contradiction.
- Equivalent wording, standard engineering synonyms, and exactly equivalent unit conversions are acceptable.
- Do not require extra reasoning or details that are not mandatory.
- Use binary scoring only. Do not output partial credit.

Return JSON only:
{"score": 0 or 1, "reason": "brief explanation"}
\end{Verbatim}
Judge prompts, panel identifiers, and run-manifest identifiers are recorded as evaluation metadata.

Three valid votes use majority decision. Two valid agreeing votes also yield a score; conflicting pairs or a single valid vote remain NA. Missing terminal candidate answers are zeros, whereas unresolved judge decisions are excluded from the scoring denominator. Judges assess semantic equivalence to the provided reference answer.

\subsection{Scoring and aggregation formulas}
For question $i$ in run $r$, each of the three judges returns a binary decision $s_{i,r}^{(j)}\in\{0,1\}$ when valid. The final response score is
\begin{equation}
\label{eq:judge_aggregation}
s_{i,r}=\begin{cases}
\operatorname{majority}\{s_{i,r}^{(j)}\}_{j=1}^{3}, & \text{three valid votes},\\
v, & \text{two valid votes, both equal to }v,\\
\mathrm{NA}, & \text{otherwise}.
\end{cases}
\end{equation}
Missing terminal candidate answers are scored as 0 under the frozen policy, whereas unresolved judge outcomes remain $\mathrm{NA}$ and are excluded from the denominator.

For domain $d$ containing $N_d$ questions, pooled domain accuracy over the three independent runs is
\begin{equation}
\label{eq:domain_accuracy}
A_d = \frac{\sum_{i\in d}\sum_{r=1}^{3} z_{i,r}s_{i,r}}{\sum_{i\in d}\sum_{r=1}^{3}z_{i,r}},
\qquad
z_{i,r}=\mathbb{I}[s_{i,r}\neq\mathrm{NA}].
\end{equation}
The primary benchmark metric is the unweighted domain average
\begin{equation}
\label{eq:domain_avg}
A_{\mathrm{avg}}=\frac{1}{3}\left(A_{\mathrm{mech}}+A_{\mathrm{arch}}+A_{\mathrm{trans}}\right).
\end{equation}
Difficulty-level results are computed analogously by restricting the sums in Eq.~\eqref{eq:domain_accuracy} to the corresponding \textit{Medium} or \textit{Hard} subset within each domain and then averaging the resulting domain accuracies with equal weight.

\section{Human Validation of the LLM Scoring Protocol}
\label{app:judge_validation}
We assess the reliability of the automatic majority-vote scorer using a blinded human review of 90 model responses, with 30 responses sampled from each of the Mechanical, Architecture, and Transportation subsets. The review set was constructed from stored R2/R3 responses using a fixed random seed. Within each domain, sampled questions were distinct, sampling did not use the LLM-judge label, and candidate model/run assignments were approximately balanced. During review, the candidate-model identity, run identifier, individual judge votes, and majority decision were hidden. The human reviewer saw the question, reference answer, and candidate response and assigned a binary correctness label under the same complete-answer criterion as the automatic scorer. Cohen's $\kappa$ is computed over binary human labels; no sampled response was marked unjudgeable.

\begin{table}[!htbp]
\centering

\caption{Blinded human validation of the LLM-based
majority-vote scoring protocol. Agreement is the
fraction of responses for which the human binary
label matches the three-judge majority decision.}
\label{tab:judge_human_validation}

\small
\renewcommand{\arraystretch}{1.12}
\setlength{\tabcolsep}{12pt}

\begin{tabular}{@{\hspace{12pt}}lccc@{\hspace{12pt}}}
\toprule

\textbf{Subset} &
\textbf{$N$} &
\textbf{Agreement (\%)} &
\textbf{Cohen's $\kappa$} \\

\midrule

Overall               & 90 & 95.6 & 0.902 \\
Mechanical            & 30 & 93.3 & 0.815 \\
Architecture          & 30 & 96.7 & 0.932 \\
Transportation        & 30 & 96.7 & 0.930 \\

\addlinespace[3pt]

Unanimous judge votes & 71 & 97.2 & 0.934 \\
2:1 split judge votes  & 19 & 89.5 & 0.791 \\

\bottomrule
\end{tabular}

\end{table}

Across the 90 reviewed responses, 86 receive the same binary decision from the human reviewer and the LLM-majority scorer. Agreement is higher when all three judges agree than when the vote is split 2:1.
\section{Run-to-Run Stability}
\label{app:run_stability}
To characterize repeated-run variability under the same primary aggregation used in the main evaluation, we compute domain-avg accuracy separately for each of the three runs. Within each run, accuracy is first computed independently on Mechanical, Architecture, and Transportation, and the three domain accuracies are then averaged with equal weight. Table~\ref{tab:run_avg_stability} reports the resulting run-wise domain-avg scores and their arithmetic mean and sample standard deviation across the three runs ($n=3$). The reported sample standard deviation summarizes run-to-run variation; results are aggregated across responses rather than by voting across runs.

\begin{table}[!htbp]
\centering
\small

\caption{Run-to-run stability under the primary domain-avg metric.
R1--R3 report domain-avg accuracy (\%) for each independent run;
the final column reports the arithmetic mean and sample
standard deviation across the three runs.}
\label{tab:run_avg_stability}

\renewcommand{\arraystretch}{1.12}
\setlength{\tabcolsep}{13pt}

\begin{tabular}{
@{\hspace{12pt}}lcccc@{\hspace{12pt}}
}
\toprule

\textbf{Model} &
\textbf{R1$\uparrow$} &
\textbf{R2$\uparrow$} &
\textbf{R3$\uparrow$} &
\textbf{Mean $\pm$ SD} \\

\midrule

\multicolumn{5}{c}{\textbf{Closed-source MLLMs}} \\
\addlinespace[1pt]

GPT-5.6 Sol &
54.15 & 53.79 & 50.02 & 52.65 $\pm$ 2.29 \\

GPT-5.6 Luna &
50.02 & 47.72 & 44.02 & 47.26 $\pm$ 3.03 \\

Qwen3.8-Max &
46.88 & 47.09 & 46.48 & 46.82 $\pm$ 0.31 \\

Gemini-3.5 Flash &
39.45 & 44.02 & 40.88 & 41.45 $\pm$ 2.34 \\

GPT-5.5 &
32.98 & 33.28 & 32.35 & 32.87 $\pm$ 0.47 \\

Qwen3.7-Flash &
25.64 & 26.77 & 24.46 & 25.63 $\pm$ 1.15 \\

Qwen3.8-Flash &
26.52 & 25.11 & 22.04 & 24.56 $\pm$ 2.29 \\

Qwen3.6-Flash &
13.68 & 12.99 & 13.32 & 13.33 $\pm$ 0.35 \\

\midrule

\multicolumn{5}{c}{\textbf{Open-weight MLLMs}} \\
\addlinespace[1pt]

Kimi-K2.5 &
23.70 & 24.99 & 25.68 & 24.79 $\pm$ 1.00 \\

Qwen3.8-27B &
22.85 & 22.66 & 21.57 & 22.36 $\pm$ 0.69 \\

Qwen3.6-27B &
22.80 & 19.88 & 18.45 & 20.38 $\pm$ 2.22 \\

\bottomrule
\end{tabular}
\end{table}

Table~\ref{tab:run_domain_stability} gives the corresponding domain-level stability statistics underlying the domain-avg scores. For each model and domain, the reported mean is computed from the three run-wise domain accuracies and matches the pooled domain accuracy in Table~\ref{tab:main_results}; the Domain Avg column is therefore identical to the final column of Table~\ref{tab:run_avg_stability}.

\begin{table}[!htbp]
\centering
\small

\caption{Domain-level run-to-run stability.
Values are mean $\pm$ sample standard deviation
of the three run-wise accuracies (\%).}
\label{tab:run_domain_stability}

\renewcommand{\arraystretch}{1.10}
\setlength{\tabcolsep}{11pt}

\begin{tabular}{@{\hspace{12pt}}lcccc@{\hspace{12pt}}}
\toprule

\textbf{Model} &
\textbf{Mechanical$\uparrow$} &
\textbf{Architecture$\uparrow$} &
\textbf{Transportation$\uparrow$} &
\textbf{Domain Avg$\uparrow$} \\

\midrule
\multicolumn{5}{c}{\textbf{Closed-source MLLMs}} \\
\addlinespace[1pt]

GPT-5.6 Sol &
58.81 $\pm$ 0.88 &
48.89 $\pm$ 4.07 &
50.26 $\pm$ 4.30 &
52.65 $\pm$ 2.29 \\

GPT-5.6 Luna &
50.38 $\pm$ 1.76 &
47.11 $\pm$ 4.29 &
44.27 $\pm$ 6.64 &
47.26 $\pm$ 3.03 \\

Qwen3.8-Max &
46.36 $\pm$ 2.02 &
48.00 $\pm$ 2.67 &
46.09 $\pm$ 1.56 &
46.82 $\pm$ 0.31 \\

Gemini-3.5 Flash &
57.47 $\pm$ 1.00 &
35.11 $\pm$ 4.29 &
31.77 $\pm$ 3.52 &
41.45 $\pm$ 2.34 \\

GPT-5.5 &
41.00 $\pm$ 1.85 &
17.78 $\pm$ 4.07 &
39.84 $\pm$ 1.35 &
32.87 $\pm$ 0.47 \\

Qwen3.7-Flash &
32.57 $\pm$ 1.20 &
16.44 $\pm$ 3.36 &
27.86 $\pm$ 1.97 &
25.63 $\pm$ 1.15 \\

Qwen3.8-Flash &
29.89 $\pm$ 0.57 &
16.44 $\pm$ 5.39 &
27.34 $\pm$ 2.07 &
24.56 $\pm$ 2.29 \\

Qwen3.6-Flash &
18.39 $\pm$ 1.52 &
6.22 $\pm$ 2.04 &
15.36 $\pm$ 1.19 &
13.33 $\pm$ 0.35 \\

\midrule
\multicolumn{5}{c}{\textbf{Open-weight MLLMs}} \\
\addlinespace[1pt]

Kimi-K2.5 &
32.18 $\pm$ 1.15 &
15.11 $\pm$ 3.36 &
27.08 $\pm$ 1.19 &
24.79 $\pm$ 1.00 \\

Qwen3.8-27B &
22.41 $\pm$ 2.63 &
20.44 $\pm$ 2.78 &
24.22 $\pm$ 1.56 &
22.36 $\pm$ 0.69 \\

Qwen3.6-27B &
31.99 $\pm$ 3.27 &
6.22 $\pm$ 2.78 &
22.92 $\pm$ 2.39 &
20.38 $\pm$ 2.22 \\

\bottomrule
\end{tabular}
\end{table}

Across the eleven reported models, domain-avg run-to-run sample standard deviations range from 0.31 to 3.03 percentage points. Some model--domain combinations, particularly in Architecture and Transportation, exhibit larger variability across runs.

\Needspace{15\baselineskip}
\section{Operational Interpretation of the Failure Stages}
\label{app:error_taxonomy}

\begin{table}[!htbp]
\centering
\small
\caption{Operational interpretation of the four conceptual failure stages used in the diagnostic analysis.}
\label{tab:error_taxonomy}
\setlength{\tabcolsep}{4pt}
\begin{tabularx}{\linewidth}{>{\raggedright\arraybackslash}p{3.9cm}Y}
\toprule
\textbf{Stage} & \textbf{Operational interpretation} \\
\midrule
Local Evidence Understanding & Relevant local text, numbers, identifiers, specifications, symbols, or geometric cues are read or interpreted incorrectly. \\
Object Association & Evidence is plausibly read but associated with the wrong engineering object, instance, space, component, annotation target, or view-specific counterpart. \\
Relation Reconstruction & Relevant objects are identified, but connectivity, adjacency, path, direction, order, continuity, scope, datum, or cross-region/view correspondence is reconstructed incorrectly. \\
Integrated Engineering Reasoning & Upstream evidence is substantially available, but the final answer fails during evidence combination, multi-step derivation, arithmetic or unit handling, aggregation, constraint application, or answer completion. \\
\bottomrule
\end{tabularx}
\end{table}

These operational categories organize the response-level error analysis reported in Section~\ref{sec:error_analysis}.

\section{Question Authoring and Quality-Control Details}
\label{app:authoring}
\paragraph{Archived authoring materials.}
Candidate construction was iterative. The archived development prompts ask a model to propose questions, candidate answers, evidence dependencies, and wording refinements from a complete engineering drawing. Engineering-background annotators review candidate questions and select the final items using drawing-grounded criteria.

\paragraph{Operational construction criteria.}
For each candidate, annotators identify a practical information need, the target engineering object or task, the drawing evidence required to answer it, the relations or inference involved, and the requirements for a complete answer. Candidates are revised or removed when grounding is unclear, the reference answer is incorrect or incomplete, wording creates unintended answer cues, engineering relevance is weak, or substantial semantic redundancy remains.

\paragraph{Difficulty distinction.}
The \textit{Medium}/\textit{Hard} distinction reflects evidence structure and reasoning demand. Medium questions typically use a bounded evidence set and relatively direct combinations or calculations. Hard questions require broader or more interdependent evidence, such as cross-region or cross-view association, spatial or topological reconstruction, ordered-dependency tracing, multi-step derivation, or joint application of multiple engineering constraints. Engineering-background experts review this distinction with reference to the source drawing and the evidence required to answer each question.

\section{Supplementary Method Study: Execution and Results}
\label{app:interventions}

\subsection{Inputs and intervention procedures}
The supplementary study compares whole-sheet direct answering with three visual-access interventions on the same 377 questions and answer requirements. Reference answers are withheld during inference, and each model--question--method combination is evaluated in three independent runs, without voting across runs or selecting the best response.

\begin{table}[!htbp]
\centering
\small

\caption{Visual-access intervention inputs and normal deployment
calls per response. Failures and fallback paths may change
the realized count.}

\renewcommand{\arraystretch}{1.12}
\setlength{\tabcolsep}{6pt}

\begin{tabularx}{0.94\linewidth}{
@{\hspace{8pt}}
>{\raggedright\arraybackslash}p{3.1cm}
>{\raggedright\arraybackslash}X
c
@{\hspace{8pt}}
}
\toprule
\textbf{Method} &
\textbf{Input / procedure} &
\textbf{Calls} \\
\midrule

Whole-sheet direct answering &
Complete drawing and question; answer directly
from the full visual input &
1 \\

Fixed overlapping crops &
Complete drawing plus four predefined overlapping
crops in one request &
1 \\

Locate then reread &
Locate at most two likely target regions, reread
the corresponding crops, and answer using the
full drawing and reread observations &
3 \\

Search then reread &
Search for multiple relevant regions, reread
up to two selected regions, and integrate
the resulting observations &
3 \\

\bottomrule
\end{tabularx}
\end{table}

Fixed overlapping crops provide the largest point estimate in Mechanical and Transportation, while search then reread performs best in Architecture and has the highest domain-avg point estimate. Locate then reread improves Architecture and Transportation over whole-sheet direct answering but is lower in Mechanical. These comparisons characterize performance under the respective visual-access input and inference procedures. Whole-sheet and fixed-crop settings normally use one model request, whereas locate/search methods normally use three, as summarized in the procedure table.

\FloatBarrier
\section{Evaluation Scope and Interpretation}
\label{app:limitations}
LED-Bench evaluates a curated set of 107 complete engineering drawing sheets and 377 questions across three disciplines. The selection emphasizes drawing-grounded evidence association, relation reconstruction, and integrated reasoning. Domain-specific drafting conventions, drawing scales, and question composition are reflected in the reported results.

Questions originating from the same sheet share a visual context. The evaluation reports pooled accuracy across three independent model runs and run-level variation; the supplementary methods study reports accuracy under its documented input and inference-call configurations. The scoring protocol evaluates answer correctness against drawing-grounded reference answers, with a separate blinded comparison between human and LLM-majority scoring decisions.

Source drawings were collected from public educational, technical, and project-document materials, while the question specifications and reference answers were constructed for LED-Bench. Release materials will document source provenance and redistribution permissions.

\end{document}